\documentclass[final]{cvpr}
\usepackage{booktabs}
\usepackage{cite}  

\usepackage{hyperref}
\usepackage{times}
\usepackage{epsfig}
\usepackage{graphicx}
\usepackage{amsmath}
\usepackage{amssymb}
\usepackage{multirow}
\usepackage{array}
\usepackage{float}
\usepackage{tikz-cd}
\usepackage[capitalize]{cleveref}

\newcommand{\R}{\mathbb{R}}
\makeatletter

\begin{document}
\title{VariGrad: A Novel Feature Vector Architecture for Geometric Deep Learning on Unregistered Data}
\author{Emmanuel Hartman$^{1}$ (elh18e@fsu.edu) \qquad Emery Pierson$^{2} $ ( emery.pierson@univ-lille.fr)  \thanks{E. Hartman was supported by NSF grant DMS-1953244.  E. Pierson has been partially supported by by the Austrian Science Fund (grant no P 35813-N). }
\\Department of Mathematics, Florida State University, Tallahassee, USA$^{1}$ \\ Institute of Mathematics, University of Vienna, Vienna, Austria $^2$ }
\maketitle

\begin{abstract}
We present a novel geometric deep learning layer that leverages the varifold gradient (VariGrad) to compute feature vector representations of 3D geometric data. These feature vectors can be used in a variety of downstream learning tasks such as classification, registration, and shape reconstruction. Our model's use of parameterization independent varifold representations of geometric data allows our model to be both trained and tested on data independent of the given sampling or parameterization. We demonstrate the efficiency, generalizability, and robustness to resampling demonstrated by the proposed VariGrad layer.
\end{abstract}
\section{Introduction}
Geometric deep learning is an exciting field that has been rapidly evolving in recent years. It involves applying deep learning techniques to problems that involve geometric data such as point clouds, meshes, graphs, and curves. The ability to effectively learn from geometric data has numerous applications in fields such as computer graphics, computer vision, and robotics. In this project, we aim to contribute to the development of geometric deep learning by proposing a novel layer for deep neural networks. Our layer leverages the gradient of the varifold norm to define a vector field on a template shape which is then seen as a feature vector defined by the input. This method has the potential to improve the accuracy of deep learning models on geometric data, which could have a significant impact on a wide range of applications. In the following article, we will describe our method in detail and present experimental results that demonstrate its effectiveness for 3D curves and shape graphs.

\subsection{Background \& Related Work}\label{ssec:background_and_related_work}

\begin{figure}[h]
    \centering
    \includegraphics[width=\linewidth]{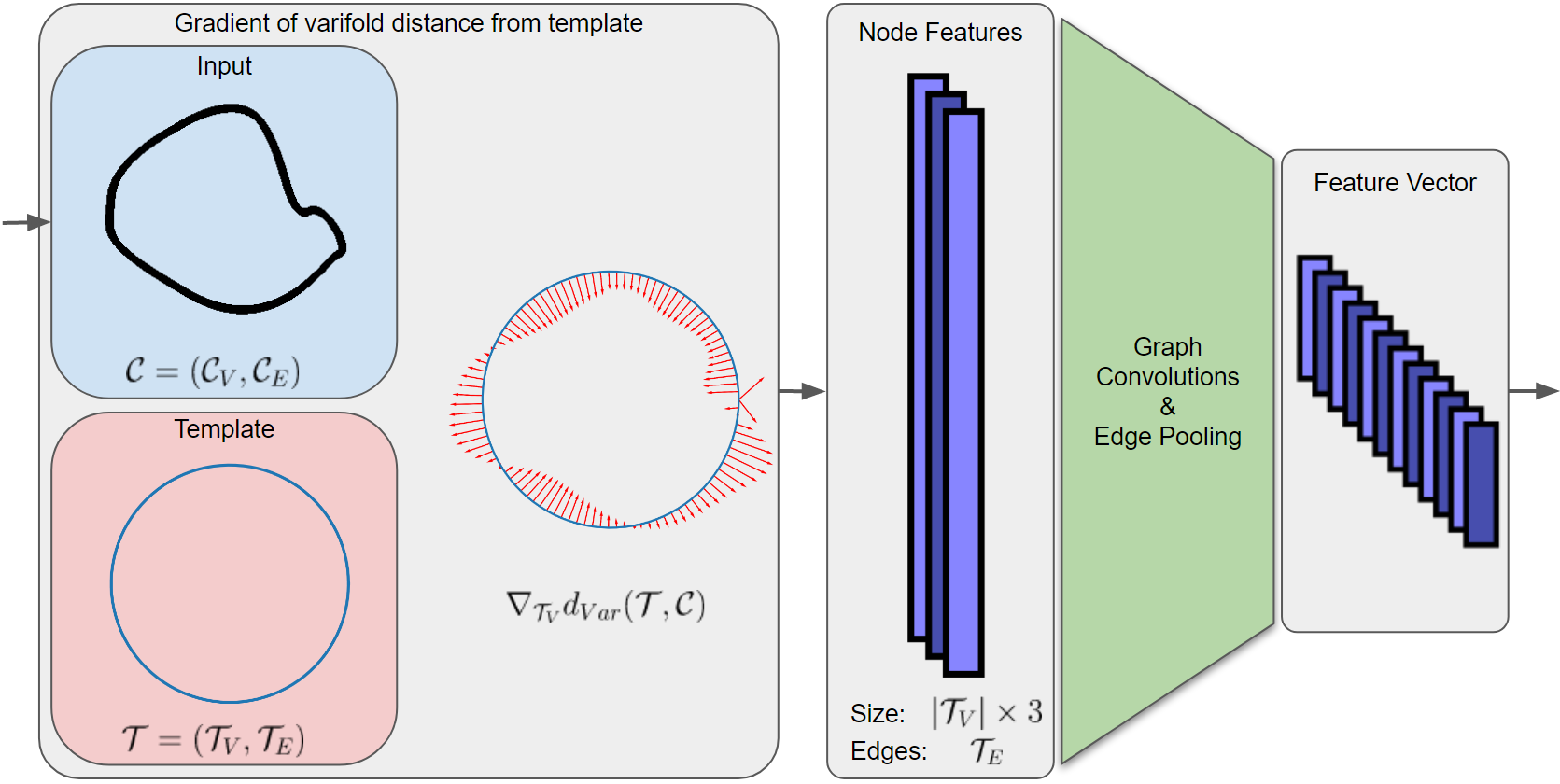}
    \caption{ Illustration of the proposed methodology. The input geometric data and template shape are transformed into varifold representations and the gradient of the varifold norm is taken with respect to the vertices of the template. This gradient is passed through graph convolutional and edge pooling layers with respect to the template edges. Resulting feature vectors can be fed into traditional deep learning frameworks for reconstruction and classification tasks.}
    \label{fig:overview}
\end{figure}

The ability to compute similarity metrics between geometric objects is crucial in many applications. However, traditional methods that rely on aligning parameterizations can be computationally expensive and may not be suitable for large or complex datasets. Varifold representations have emerged as a powerful tool for computing parameterization invariant similarity terms for geometric objects. Unlike traditional approaches, varifold representations are independent of the parameterization of the data, making them well-suited for cases where objects do not share a point-to-point correspondence. The space of varifolds has been equipped with a reproducing kernel Hilbert space (RKHS) norm. The benefits of using this norm are that it is robust to noisy data and can be efficiently computed. The RKHS norm is also differentiable, making it suitable as a loss function for optimization-based shape matching~\cite{sukurdeep2019, hartman2023elastic}, or deep learning models for shape reconstruction~\cite{BESNIER2023}.  

Most of the pioneering works use well-defined shape descriptors to obtain a suitable representation of curves or shape graphs. Indeed, it is theoretically possible to retrieve a curve from its geometric moments, or its affine integral signature~\cite{feng2010classification}, that have been applied in the context of curve classification. However, retrieving curves from such features implies solving costly optimization problems~\cite{kousholt2021reconstruction}. The Square Root Velocity Field framework has been used in elastic matching~\cite{srivastava2010shape} on 3D 
curves. It has been used on face-extracted curves for face identification~\cite{veltkamp2011shrec} and on extremal human curves~\cite{slama_3d_2014} for human pose retrieval. However, this elastic matching approach requires solving the problem of parameterization for each pair of curves. Moreover, their extension to shape graphs remains an active area of research~\cite{guo2021quotient,guo2022statistical,sukurdeep2022new}.

In the recent years, a few works have shown the superiority of deep learning-based methods over the previous approaches. However, some approaches require strong hypotheses on the data, such as a common parameterization~\cite{hartman2021supervised}, or an ordering of the curve or shape graph\cite{ha2018a, xu2021multigraph}. Meanwhile, a few approaches using convolutional neural networks on binary images have been proposed for sketch representation~\cite{qin2022shrec, manda2022sketchcleannet}. Not only those approaches don't take profit from the geometric information of curves, but they also do not extend to 3D data. Moreover, most of those approaches require training on large datasets~\cite{sketchyx, ha2018a} to be efficient.

Geometric deep learning~\cite{bronstein2017geometric} tries to circumvent those problems compared to the previous approaches. 
In particular, a few methods address the problem of dealing with data that lacks point-to-point correspondences or fixed shapes.
Those methods interpret point clouds with an arbitrary number of inputs as a feature vector of a single-shaped object. PointNet~\cite{Charles_PointNet_2017} is a notable example of such methods, designed specifically for processing and classifying point clouds. PointNet employs a shared multi-layer perceptron (MLP) network to transform individual points into a high-dimensional feature space. The feature vectors are then aggregated using a symmetric function, such as max pooling, to generate a fixed-length representation of the entire point cloud. This representation can be fed into a fully connected neural network for classification or other downstream tasks. To overcome the limitation of solely relying on point cloud information without considering connectivity and local relationships between points, the Dynamic Graph Convolutional Neural Network (DGCNN)~\cite{wang2019dynamic} method was introduced. DGCNN utilizes k-nearest neighbors (kNN) to establish edges between points, enabling the construction of a graph representation. By employing graph convolutional and edge pooling layers, DGCNN generates high-dimensional feature vectors for each point and aggregates them using a symmetric function. This results in a fixed-length representation of the entire point cloud that depends on the local relationships between points, which can be fed into a fully connected neural network for classification or other tasks. While some other architectures~\cite{mitchel2021field, sharp2022diffusionnet, Wiersma2022DeltaConv} have shown more expressivity on 3D point clouds and surfaces, they rely on surface-specific operators such as the Laplace-Beltrami operator~\cite{sharp2022diffusionnet}, which cannot be used for 3D curves or shape graphs.

\subsection{Contributions}\label{ssec:contributions}
In contrast, our proposed method directly operates on curves and shape graphs rather than their vertices as point clouds. We introduce a novel approach that leverages varifold representations to construct consistent-sized feature vectors, enabling the representation of geometric data irrespective of its sampling or registration. By utilizing varifold representations, our method captures the intrinsic properties of curves and shape graphs, leading to enhanced representations for geometric data. We propose a novel solution to this problem that leverages the differentiability of the varifold similarity metric. We compute the gradient of the distance between a template shape and the input shapes with respect to the points of the template shape.  The resulting gradient is  a vector field on the template shape that describes the optimal direction of deformation of the vertices of the template to minimize the distance between the shapes. This vector field can be interpreted as a feature vector for each input shape whose dimension is fixed regardless of the discretization or parameterization of the input shape. Traditional deep learning layers can process this feature vector, which is represented as a vector field lying on our template shape. Additionally, the vector field can be convolved with respect to the geometry of our template shape. For open curves, this is a 1D convolution, and for closed curves, this convolution is performed with cyclic padding. For surfaces and shape graphs, a graph convolution can be performed with respect to the underlying connectivity of the template shape. This approach provides a robust and flexible method for converting unregistered geometric objects into a fixed-dimensional feature vector that can be utilized in traditional and geometric deep-learning models.


\section{Feature Vectors for 3D Curves and Shape Graphs}
We focus on discrete 3D curves and shape graphs which are composed of vertices $V\subseteq\R^3$ and linear edges $E$ connecting these vertices. However, a major challenge in working with such geometric data is the reliance on specific parameterizations, which can make downstream learning tasks and analysis difficult. To address this challenge, we adopt varifold representations as a powerful tool for characterizing and studying geometric objects without being dependent on their parameterizations.
\subsection{Varifold Representations of 3D Curves and Shape Graphs} 
\begin{figure*}[h]
    \centering
    \includegraphics[width=\linewidth]{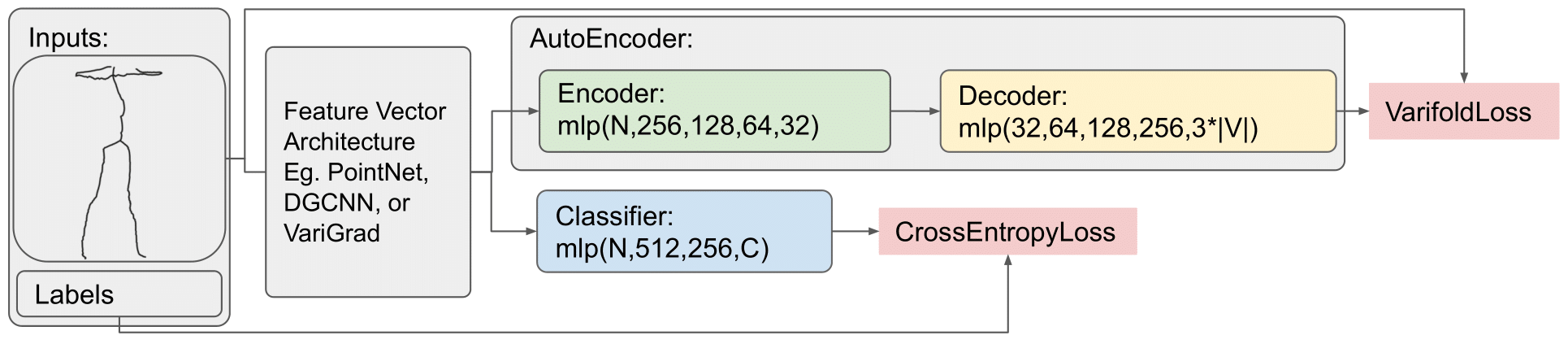}
    \caption{Network Architectures for Downstream Tasks: A diagram of the classifier and auto-encoder networks used for numerical experiments. Unparameterized 3D curves or shape graphs are given as inputs and feature vectors are extracted via a feature vector architecture. These feature vector architectures are then combined with the downstream architectures described in this figure and the combined networks are trained for their respective tasks. }
    \label{fig:downstream}
\end{figure*}\label{sec:intrVari} In recent years, varifold representations of geometric data have seen many applications in the field of shape analysis \cite{sukurdeep2019, Kaltenmark_2017_CVPR, charon2013,CHARON2020441}. The space of varifolds is given by Borel measures on the Cartesian product of $\R^3$ and the unit sphere $S^2$ denoted by $\mathcal{M}(\R^3\times S^2)$. Each discrete shape graph, $\mathcal{C}=(V,E)$, is mapped to a finitely supported measure $\mu\in\mathcal{M}(\R^3\times S^2)$ where each support corresponds to each edge and is given by the centroid and normalized tangent vector of the edge. The mass of the support associated with an edge in the varifold representation is determined by its length. This mapping ensures that regardless of the specific parameterization or ordering of points and edges in a shape graph, it will be identified with the same varifold representation. The space of varifolds is equipped with a Reproducing Kernel Hilbert Space (RKHS) norm, we define a suitable kernel function that measures the similarity between varifolds. The kernel function captures the inner product structure in the space of varifolds, allowing us to compute distances and perform various operations. This kernel function can be defined coordinate-wise by taking the product of a Gaussian kernel on $\R^3$ and a Binet kernel on $S^2$. The inner product defined by this kernel is given by
\[\langle\mu,\nu\rangle_{Var}=\int_{\R^3\times S^2}\int_{\R^3\times S^2} e^{-a\|x_1-x_2\|^2} \langle u_1,u_2\rangle^2 d\mu \, d\nu\]
where $a$ is a balancing parameter. From this inner product, a  distance on the space of varifolds is given by
\[\|\mu-\nu\|^2=\langle\mu,\mu\rangle_{Var}+\langle\nu,\nu\rangle_{Var}-2\langle\mu,\nu\rangle_{Var}.\]
We define the dissimilarity between two discrete shape graphs $\mathcal{C}_1=(V_1,E_1)$ and $\mathcal{C}_2=(V_2,E_2)$ as the varifold distance between the associated finitely supported varifolds $\mu_1,\mu_2$ respectively. In particular, we write the dissimilarity between two shape graphs as $d_{Var}(\mathcal{C}_1,\mathcal{C}_2)=\|\mu_1-\mu_2\|$. As these varifolds are finitely supported the inner product can be approximated by a sum over $E_1$ and $E_2$ given by 

\[\langle\mu_1,\mu_2\rangle_{Var}=\sum_{e_i\in E_1}\sum_{e_j\in E_2} e^{-a\|c_i-c_j\|^2} \langle u_i,u_j\rangle^2 l_i l_j.\]
Here $c_i, c_j$ denote the centroids of $e_i, e_j$, $u_i,u_j$ denote the normalized tangent vectors of $e_i,e_j$, and $l_i,l_j$ denote the lengths of $e_i,e_j$. 

By utilizing varifold representations equipped with a RKHS norm, we can effectively and efficiently capture the essential geometric information of 3D curves and shape graphs in a parameterization-independent manner. This allows for meaningful and efficient comparisons, analysis, and processing of geometric data, as well as facilitating tasks such as shape matching, recognition, and reconstruction. Moreover, the invariance of varifold representations to parameterization changes provides robustness and consistency in geometric deep learning tasks.

\subsection{VariGrad Feature Vector Architecture}\label{sec:VariGrad}

Our architecture is designed for data-driven applications where the geometric data used for training or applying the model can be matched to a template shape $\mathcal{T}$. Leveraging this template shape, we compute the gradient of the varifold distance between each input shape and the template shape, specifically with respect to the vertices of the template shape. 
This computation results in a vector field on the template that can be viewed as a representation of the input. This data is fed into a sequence of convolutional layers where the convolution is performed with respect to the template shape~\cite{kipf2017semisupervised}. The resulting vector field is concatenated into a fixed-dimension representation of the input data regardless of the sampling of the input data.  A visual representation of the VariGrad architecture is given in Figure~\ref{fig:overview}. Moreover, we have made a Pytorch implementation of this framework as well as the trained VariGrad models available on Github\footnote{\url{https://github.com/emmanuel-hartman/Pytorch\_VariGrad}}.

Our framework sees improved results compared to methods that combine pointwise Multi-Layer Perceptron (MLP) and feature point aggregation. This is mainly due to the utilization of the geometric prior given by the template shape on the input data. Moreover, rather than treating the points individually our method is able to efficiently process whole curves and shape graphs even in the unregistered setting, thanks to their varifold representations. Moreover, as they use MLP to embed each point in the point cloud in a high dimensional space, PointNet and DGCNN require more trainable parameters than the graph edge convolutions~\cite{kipf2017semisupervised} used by VariGrad. 

\section{Experimental Results}\label{sec:numerical_experiments}
The following experiments aim to compare the performance of the proposed VariGrad layer to the commonly used PointNet and DGCNN architectures in geometric deep learning applications where point-to-point correspondences are unknown. Each experiment constructs  models that utilize the PointNet, DGCNN, and VariGrad architectures respectively to learn a consistent size feature vector. The feature vectors from these architecturesare used in further learning tasks such as classification or shape reconstruction. Other than the differences in resources required by the different feature vector architectures these models use the same trainable parameters, loss functions, optimization parameters, and number of training epochs.

\subsection{Datasets}\label{ssec:datasets}
\begin{figure}[h]
    \centering
    \includegraphics[width=\linewidth]{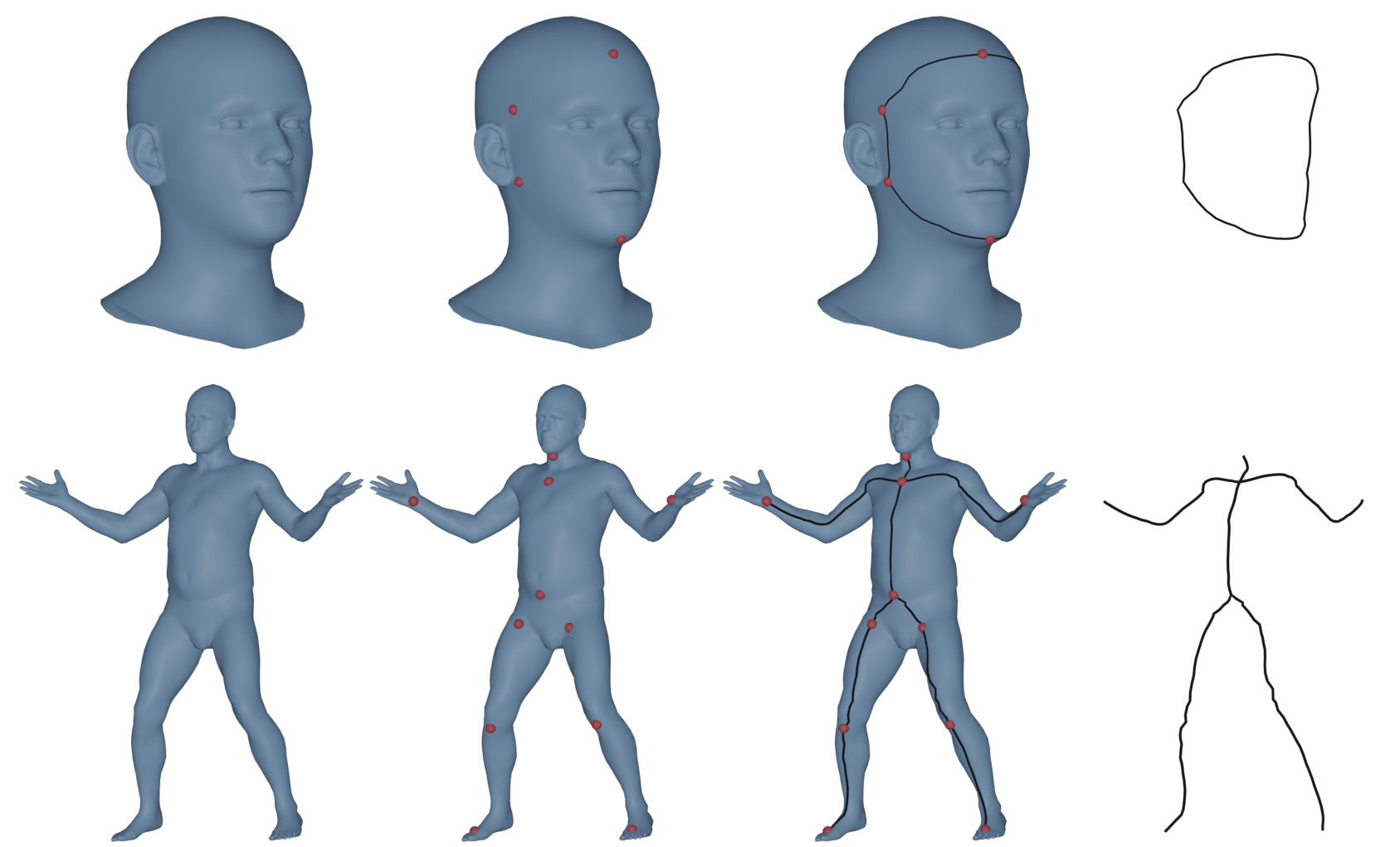}
    \caption{Examples of curve and shape graph extraction used for the creation of the example datasets. Surface meshes are given as inputs, landmarks are identified, geodesics along the surface are found between the landmark points, and finally these geodesic curves are concatenated to create the curves and shape graphs. }
    \label{fig:extraction}
\end{figure}

To evaluate the performance of the methods, similar networks will be trained using these layers on two different datasets. We consider one dataset of 3D curves and one dataset of 3D shape graphs that are extracted from 3D surface data. We first extract landmarks from the shapes available in those datasets and compute the surface geodesics on the given shapes, giving curves that represent the pose of the 3D shapes. An illustration of the process is visualized in~\Cref{fig:extraction}. We use this geometric data as input of the different models. In both datasets, the extraction process produces unregistered curves and shape graphs, meaning that there is no prior point correspondence present in the data. For both of the datasets we consider, we perform a random split between training and testing set, ensuring that the classes (face or body identities) are well-balanced between the two splits. Moreover, we take the template shape to be a random shape from the corresponding dataset.
The first dataset comprises boundary curves of human faces extracted from the COMA dataset of human face motions \cite{COMA:ECCV18}. The extraction process is performed on all 20465 elements of the COMA dataset, resulting in 20465 closed curves in three dimensions. These curves are unparameterized and have a variety of samplings due to the extraction process with numbers of vertices ranging from 64 to 96. This dataset is then separated into a training set of 18506 curves and a validation set of 1959 curves.  Along with each curve, a label indicating which of the eight identities present in the COMA dataset from which the curve was extracted. The second dataset consists of 41220 shape graphs representing human pose, extracted from the Dynamic FAUST dataset of 4D human motion scans\cite{dfaust:CVPR:2017}. Again these shape graphs are unparameterized where the numbers of vertices in the shape graphs range from 111 to 356. Similar to the dataset of curves, the data is split into 35000 training samples and 6220 testing samples and an identity label is included for each shape graph in the dataset, corresponding to one of the twelve identities. By training and comparing the performance of the networks on these datasets, insights can be gained regarding the suitability and effectiveness of the VariGrad layer when compared to the state-of-the-art PointNet and DGCNN methods.

\subsection{Classification}\label{ssec:classification}
We evaluate the classification models based on their classification accuracy on the testing data. The results show that the model utilizing the VariGrad architecture outperforms the PointNet and DGCNN architectures. This difference in performance is likely due to the sparser set of points provided by the vertices of the three-dimensional curves considered in this experiment. The lower number of points may not be sufficient for the PointNet and DGCNN architectures to learn representative feature vectors. The downstream network is an MLP, for which the architecture is outlined in Figure \ref{fig:downstream}. Here $N$ denotes the dimension of the feature vector recovered by the respective architectures and $C$ denotes the number of classes in the data set. The input and hidden fully connected layers are followed by a ReLU activation function and a batch normalization layer. 
\begin{table}[h]
    \centering
    \footnotesize
    \aboverulesep=0ex
    \belowrulesep=0ex
    \begin{tabular}{c|cc|cc}
    \multicolumn{1}{c|}{} & \multicolumn{2}{c|}{Face Curve Data} & \multicolumn{2}{c}{Shape Graph Data} \\\cmidrule{2-5}
         &Classification& Train Time &Classification&Train Time  \\
         &Accuracy&(s\,/\,batch)&Accuracy&(s\,/\,batch)  \\\hline
         PointNet&29.36&0.157&78.08&0.148\\
         DGCNN&62.71&0.099&\textbf{83.29}&0.113\\
         VariGrad&\textbf{84.37}&\textbf{0.045}&\textbf{83.31}&\textbf{0.046}\\
    \end{tabular}
    \caption{Classifications Results: We present the results of the classification network utilizing three different feature vector architectures. We present the percentage of correct identity classifications by our model on an unseen testing set. Further, we present the mean training time per batch (10 samples) of each network to underscore the efficiency of the VariGrad approach. }
    \label{tab:clasification}
\end{table}
We train models using each feature vector architecture on both the COMA curve training set and DFAUST shape graph training set. We report the results of these experiments in Table~\ref{tab:clasification} along with the average training time of each model. In both the cases of curves and shape graphs, the models using the VariGrad feature vector architecture produce higher classification accuracies than the models utilizing the PointNet or DGCNN architectures. Moreover, the reliance on less trainable parameters means that the average training time of the VariGrad architecture is significantly lower than that of PointNet and DGCNN.

\subsection{Shape Reconstruction}\label{ssec:Shape_Reconstruction}
\begin{figure*}[!ht]
    \centering
    \includegraphics[width=.32\linewidth]{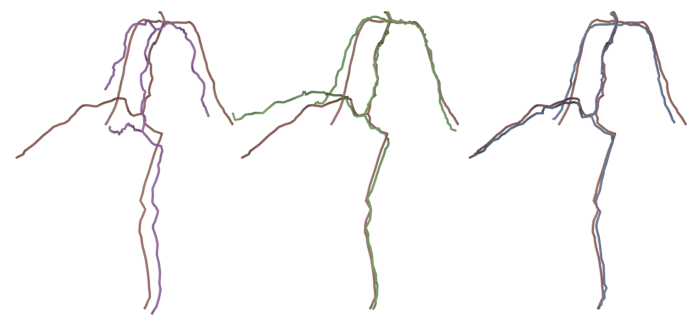}\,\vline\,
    \includegraphics[width=.32\linewidth]{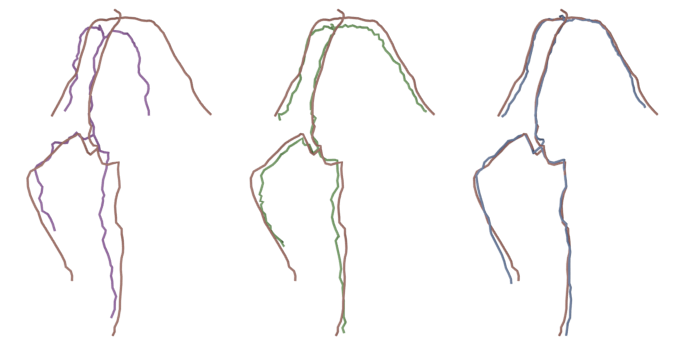}\,\vline\,
    \includegraphics[width=.32\linewidth]{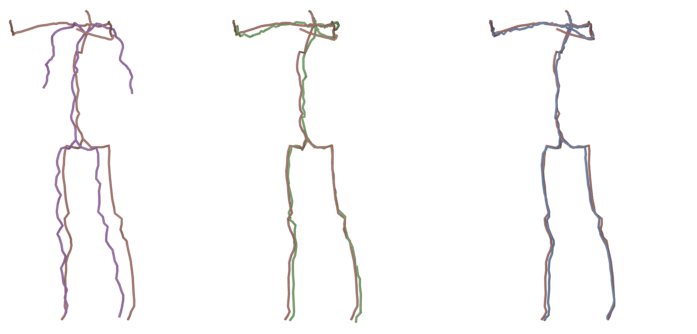}
    \caption{Qualitative comparison of shape reconstruction results for three examples from the testing set of the human pose shape graph dataset. We display the target shape graph (orange), with the reconstructions of PointNet (purple), DGCNN (green), and VariGrad (blue).  }
    \label{fig:reconstruction}
\end{figure*}
For the task of shape reconstruction, three fully connected autoencoder models are trained to learn a latent code representation of the shapes based on the feature vectors produced by PointNet, DGCNN, and VariGrad, respectively. The auto-encoder architecture is outlined in Figure \ref{fig:downstream} where $N$ denotes the dimension of the feature vector recovered by the respective architectures and $|V|$ denotes the number of desired vertices for the shape reconstruction.
\begin{table}[h]
    \centering
    \footnotesize
    \aboverulesep=0ex
    \belowrulesep=0ex
    \begin{tabular}{c|cc|cc}
    \multicolumn{1}{c|}{} & \multicolumn{2}{c|}{Face Curve Data} & \multicolumn{2}{c}{Shape Graph Data} \\\cmidrule{2-5}
         &Mean& Train Time &Mean&Train Time  \\
         &Error&(s\,/\,batch)&Error&(s\,/\,batch)  \\\hline
         PointNet&0.0053&0.149&0.0039&0.174\\
         DGCNN&0.0010&0.113&0.0015&0.133\\
         VariGrad&\textbf{0.0004}&\textbf{0.073}&\textbf{0.0005}&\textbf{0.072}\\
    \end{tabular}
   \caption{Shape Reconstruction Results: We present the results of the shape auto-encoder network utilizing three different feature vector architectures. We present the mean varifold error for an unseen testing set as well a the mean training time per batch (10 samples).}
    \label{tab:reconstruction}
\end{table}
Again, we train models using each feature vector architecture on both the COMA curve training set and DFAUST shape graph training set. To evaluate each network we compute the mean of the squared varifold norm between unseen testing sets of shapes and the reconstructions produced by each network. We report the results of these experiments in Table~\ref{tab:reconstruction} with the average training times of each model. Again, the VariGrad architecture requires less training time than PointNet or DGCNN. Meanwhile, the VariGrad architecture produces a lower mean error than PointNet or DGCNN for both curves and shape graphs. For the testing set of shape graphs, we show a qualitative comparison of the reconstructions produced by each of the feature vector architectures in Figure~\ref{fig:reconstruction}. This comparison confirms that VariGrad not only produces numerically superior results but also visually more accurate reconstructions of 3D geometric data.

\subsection{Generalizability and Reparameterization Invariance}
The final set of experiments we present demonstrates the generalizability of our model and the invariance of our model to reparameterizations of the input data. To do this, we consider two additional testing sets of shape graphs. First, we consider resampled versions of the DFAUST shape graph testing set. For this dataset, we apply random reparameterizations to each curve in the shape graph. This includes identity labels so that we can evaluate both the reconstruction and classification networks on the data.  Second, we extract 90 shape graphs from the FAUST dataset of 3D scans of human poses. This dataset includes less extreme poses than the DFAUST shape graph dataset but does include some variability in the rigid alignment of the shape graphs not seen in the training data. 
\begin{table}[h]
    \centering
    \footnotesize
    \aboverulesep=0ex
    \belowrulesep=0ex
    \begin{tabular}{c|cc|c}
        \multicolumn{1}{c|}{} & \multicolumn{2}{c|}{Reparameterized DFAUST} & \multicolumn{1}{c}{FAUST} \\\cmidrule{2-4}
         &Classification&Reconstruction&Reconstruction\\
         &Accuracy&Mean Error&Mean Error\\\hline
         PointNet&49.98&0.0040&0.0061\\
         DGCNN&50.63&0.0049&0.0064\\
         VariGrad&\textbf{80.91}&\textbf{0.0005}&\textbf{0.0040}\\
    \end{tabular}
   \caption{Generalizability Results: Here we test our pre-trained shape graph reconstruction and classification models on two additional datasets which include variability unseen in the training. }
    \label{tab:generalizability}
\end{table}
\begin{figure}[h]
    \centering
    \includegraphics[width=\linewidth]{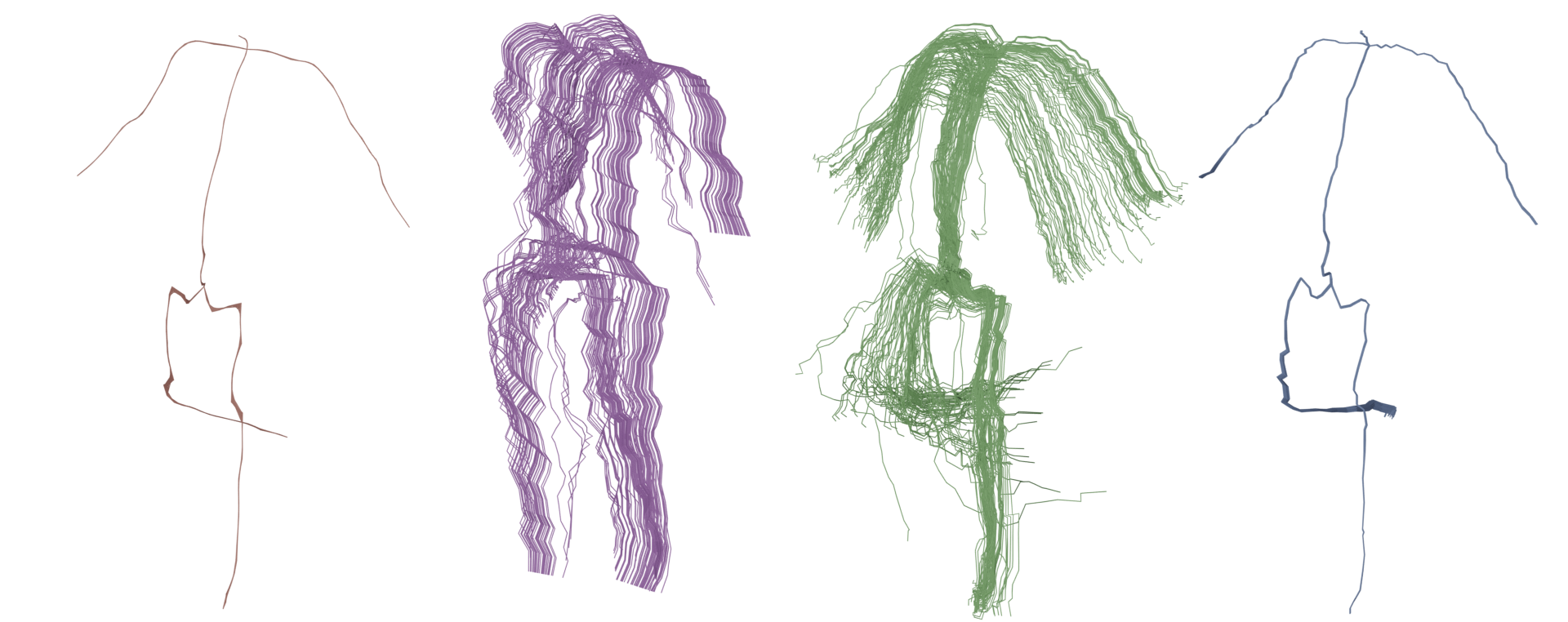}
    \caption{Qualitative comparison of each model's invariance to reparameterization. We display 100 resampled versions of a single shape graph (orange) and the 100 reconstructions produced by PointNet (purple), DGCNN (grean), and VariGrad (blue).  }
    \label{fig:invariance}
\end{figure}
In Table~\ref{tab:generalizability}, we report the evaluation metrics for the trained classification and reconstruction models applied to the reparameterized DFAUST shape graphs and for the trained reconstruction model applied to the FAUST shape graphs. In these results, we show that the VariGrad architecture produces superior results for both classification and reconstruction tasks. In particular, for the reparameterized DFAUST data, we report similar results to the original testing set in both classification and reconstruction tasks. Meanwhile, the DGCNN architecture sees a significant decrease in classification and reconstruction accuracy for the reparameterized data. 

Finally, we propose an experiment to evaluate qualitatively the parameterization invariance of the different models. We created 100 reparameterizations of a single randomly selected shape graph from the DFAUST shape graph data set and applied them as input for the trained autoencoders. We collected the 100 outputs for each method and display them in~\Cref{fig:invariance}. We observe that our model keeps a very low variation of output shapes, as opposed to both PointNet and DGCNN. This visually confirms the invariance of our model to resampling and reparameterization.

\section{Conclusion}\label{sec:conclusion}
In this paper, we presented VariGrad, a new feature vector architecture for learning tasks on geometric data such as curves and shape graphs. As we demonstrated in the experiments, by leveraging the varifold representation, this architecture produces superior results in downstream learning tasks. Moreover, varifold representations are not dependent on the given parameterization of the data and thus the VariGrad layer is robust to reparameterization and resampling of geometric data. We plan in the short future to extend this work to new applications, such as hand sketches. However, as the topology of hand sketches often vary in a same class of drawnings, the choice of Varigrad's template shape would have to be done more carefully. Moreover, as the varifold representation of curves has a natural extension to surfaces and point clouds, we plan to focus on extending VariGrad to such objects in the future.

\newcommand{\etalchar}[1]{$^{#1}$}

\end{document}